\documentclass{article}
\usepackage{spconf,amsmath,graphicx}
\usepackage{algorithm2e}
\usepackage{booktabs}
\usepackage{bbding}

\title{Aerial View River Landform Video segmentation: A Weakly Supervised Context-aware Temporal Consistency Distillation Approach}
\name{Chi-Han Chen\textsuperscript{1}, Chieh-Ming Chen\textsuperscript{2}, Wen-Huang Cheng\textsuperscript{3} Ching-Chun Huang\textsuperscript{1} \sthanks{ This work was financially supported in part by the National Science and Technology Council, Taiwan, under Grant NSTC-112-2221-E-A49-089-MY3, Grant NSTC-110-2221-E-A49-066-MY3, Grant NSTC-111- 2634-F-A49-010, Grant NSTC-112-2425-H-A49-001, NSTC-112-2628-E-002-033-MY4, NSTC-112-2634-F-002-002-MBK and in part by the Higher Education Sprout Project of the National Yang Ming Chiao Tung University and the Ministry of Education (MOE), Ministry of Digital Affairs (MODA) and Environmental Protection Bureau, Taichung City Government, Taiwan.}}
\address{National Yang Ming Chiao Tung University\textsuperscript{1}, Overseas Chinese University\textsuperscript{2}, National Taiwan University\textsuperscript{3}}
\begin{document}
%
\maketitle
\begin{abstract}

The study of terrain and landform classification through UAV remote sensing diverges significantly from ground vehicle patrol tasks. Besides grappling with the complexity of data annotation and ensuring temporal consistency, it also confronts the scarcity of relevant data and the limitations imposed by the effective range of many technologies. This research substantiates that, in aerial positioning tasks, both the mean Intersection over Union (mIoU) and temporal consistency (TC) metrics are of paramount importance. It is demonstrated that fully labeled data is not the optimal choice, as selecting only key data lacks the enhancement in TC, leading to failures. Hence, a teacher-student architecture, coupled with key frame selection and key frame updating algorithms, is proposed. This framework successfully performs weakly supervised learning and TC knowledge distillation, overcoming the deficiencies of traditional TC training in aerial tasks. The experimental results reveal that our method utilizing merely 30\% of labeled data, concurrently elevates mIoU and temporal consistency ensuring stable localization of terrain objects. Result demo : https://gitlab.com/prophet.ai.inc/drone-based-riverbed-inspection
\end{abstract}
\section{Introduction}
\label{sec:intro}

River landform analysis is integral to environmental engineering \cite{fryirs2022assemblages}, finding broad applications in hydrological and hydraulic analysis, sediment transport modeling \cite{zhang2022warming}, sedimentation pattern analysis, river habitat restoration \cite{serra2022restoring}, and river channel management \cite{liro2022first}. To comprehensively analyze these aspects, understanding the spatial distribution of river channels, vegetation cover, exposed areas, gravel and sand source erosion zones, among other factors, is essential. With recent advancements in object segmentation technology enabling real-time execution speeds, the use of unmanned aerial vehicles (UAVs) for visual sensing has gained widespread traction across various remote sensing and monitoring applications \cite{he2022manet}. Imagery analysis of vegetation coverage, exposed ground, and sediment accumulation locations within riverbeds necessitates capture during periods of river drought. However, publicly available datasets of this nature are exceedingly scarce. Furthermore, as illustrated in figure 1, the morphology and boundaries of river landscapes during these periods are notably irregular.Additionally, we have found that many deep learning models lose their effectiveness when applied to aerial imagery.
This paper introduces an innovative model architecture tailored to address the challenges associated with segmenting river channels, vegetation, exposed land and sedimentation using UAV aerial images.

\begin{figure}[t]
    \centering
    \includegraphics[width=0.9\linewidth]{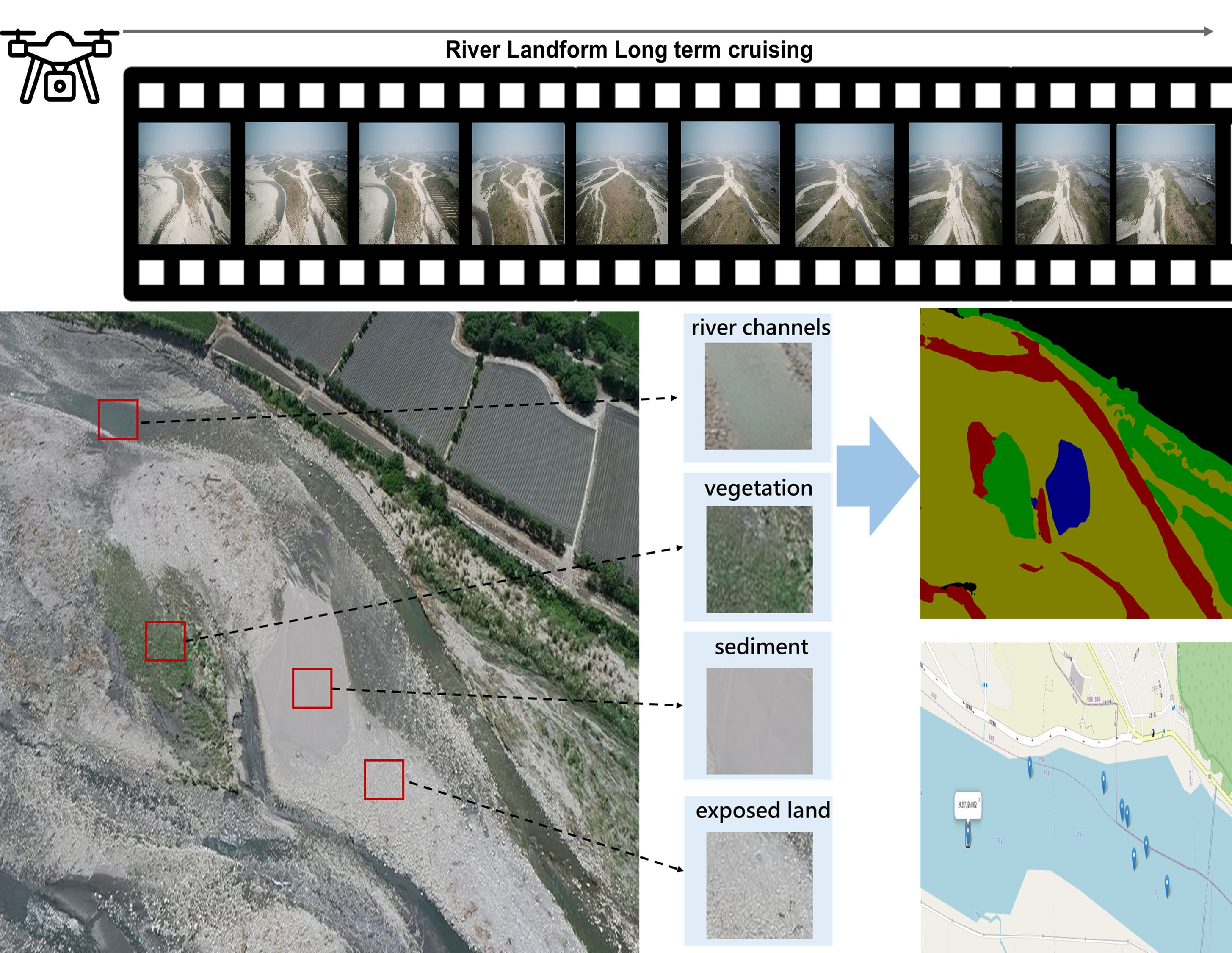}
    \caption{This long-term video dataset is captured during the dry season of the Dajia River and Da'an River in Taiwan. To accurately locate large areas of sediment and exposed ground after segmentation, the use of segmentation techniques becomes essential, thereby giving rise to the need for applications that involve temporal consistency and weakly supervised learning.}
    \label{fig:banner}
\end{figure}

Segmentation \cite{waqas2019isaid}, offering pixel-wise inference, provides fine-grained quality and expanded application possibilities compared to object detection. However, it comes with the formidable task of manual annotation. In UAV remote sensing, challenges like intensive manual annotation \cite{kuo2011unsupervised}, moving object detection, and maintaining frame consistency are significant. To address these problem, few-shot segmentation \cite{li2021adaptive} may helps reduce manual data but focuses mainly on matching support and query images, often neglecting temporal frame consistency. Also, limited terrain segmentation datasets hinder prototypical feature extraction in learning. On the other hand, traditional approached \cite{varghese2021unsupervised} for temporal consistency largely relies on optical flow algorithms to calculate temporal loss. However, these algorithms become ineffective at high altitudes far from the ground. Such shortcomings make it difficult for segmentation to maintain consistency over time in drone vision applications, resulting in an increased rate of false detection in remote sensing applications.


Consequently, we pose the question: {\bf Is it possible, in the absence of relevant public datasets and in domains where optical flow applications prove ineffective, to train segmentation models on a limited selection of key frames within aerial monitoring imagery while still maintaining temporal consistency?}

To achieve the aforementioned objectives, we introduce a teacher-student architecture that leverages Video Object Segmentation (VOS) in lieu of optical flow algorithms to expand application flexibility within the temporal domain. Furthermore, we propose algorithms for key frame selection and key frame updates. The key frame selection algorithm enables weakly supervised learning using sparse annotated data, while the key frame update algorithm governs the mechanism by which VOS assigns memory frames, thus modifying VOS to conform to the definitions of temporal consistency loss calculation.


Our primary contributions encompass:
Application of image segmentation technology to river landform analysis tasks, facilitating the identification of exposed areas and sediment in rivers.
We combined VOS and segmentation models along with key frame selection and key frame update to achieve temporal consistency in weakly supervised training.
The experimental results show that within our training framework, even with training on few labeled data, the model's mIoU and TC still surpass those of models trained on fully labeled data.

\begin{figure*}[ht]
    \centering
    \includegraphics[width=0.9\linewidth]{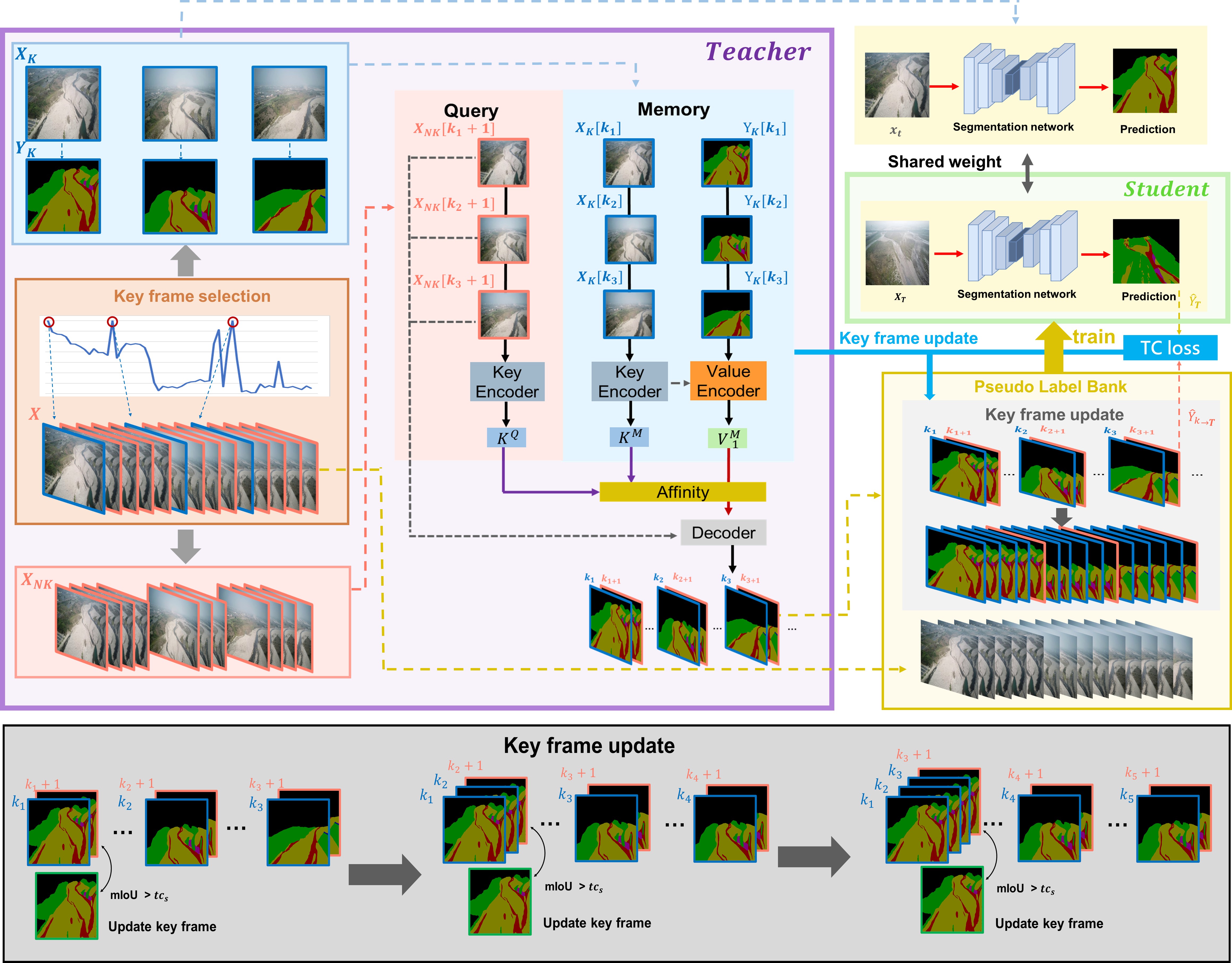}
    \caption{The main architecture of this study includes a TeacherNet and a StudentNet. The Teacher transfers temporally consistent knowledge to the student through a Pseudo Label Bank. The addition of Key frame selection aims to achieve weakly supervised learning. Key frame updates are based on temporal consistency, updating stable results into the key frame set, ensuring that frame t-1 is included in the key frames in subsequent iterations.}
    \label{fig:main_nethod}
\end{figure*}

\section{Related works}
\label{sec:format}

\subsection{Aerial Remote Sensing}
\label{ssec:Aerial_remote_sensing}

In recent research, UAV has been utilized as a carrier for aerial dynamic IoT \cite{chakravarthy2022dronesegnet}, and as one of the primary tools in various remote sensing applications \cite{liu2021light}. However, compared to conventional ground-based visual perception applications, the visual perception technology for aerial vision, involving scale variation, moving objects and the preservation of temporal consistency, presents significant challenges \cite{siam2022temporal}. There is also a considerable need for manual data labeling, which is compounded by a relative shortage of open datasets. This scarcity restricts the effectiveness of using transfer learning for fine-tuning different tasks. 

\subsection{Temporal Consistency of Segmentation}
\label{ssec:TC_Seg}

In numerous studies, it has been noted that current segmentation models in training predominantly optimize based on the metric mIoU. However, in real-world applications of unmanned vehicles, the input image frames are extracted from video streams\cite{chang2013rectangling}. Given that segmentation models are optimized solely with mIoU as the objective, they may fail to adequately learn temporal characteristics. To address this issue, Varghese et al. \cite{varghese2021unsupervised} and colleagues have proposed the criterion of temporal consistency for video segmentation, particularly in the context of autonomous driving applications. This concept primarily relies on estimating the current frame's segmentation based on the previous frame's result and the motion between these frames (e.g., using optical flow techniques\cite{liu2020efficient}). However, the reliance on optical flow for motion estimation is contingent on the number of feature points in the input images. In UAV views, insufficient or erroneous feature points could lead to the segmentation model learning noise instead of valid information.

\subsection{Video Object Segmentation}
\label{ssec:VOS}


Video Object Segmentation (VOS) in computer vision is a task that involves identifying and tracking objects within video sequences. It can be categorized into appearance-based methods \cite{chen2022video} and appearance-motion based methods \cite{cho2023treating}. The latter places greater emphasis on understanding the motion and changes of objects over time. This approach primarily utilizes cross-attention between support and query images for mask inference in query images. It necessitates that the object to be tracked appears in the reference frame, often the first frame. Most VOS model training is still based predominantly on full-shot training, while \cite{yan2023two} firstly  propose training the entire sequence with just two frames, reducing the labeling workload significantly. However, it struggles to segment objects not present in the reference frame when there is a significant disparity between the two.

\section{Method}
\label{sec:pagestyle}


In our study, we initially outline the following characteristics for the task of River Landform Analysis using UAVs:

1. To reduce the amount of labeled data, we must train with a small dataset. However, to maintain detection accuracy during flight, we must ensure the temporal consistency of segmentation.

2. To apply temporal consistency, we must find methods that are applicable to aerial imagery and modify them to fit the architecture of temporal consistency loss.

To address these challenges, we propose an architecture as illustrated in Figure \ref{fig:main_nethod}. This architecture encompasses several key components: Key Frame Selection, Model Initialization, Temporal Consistent Learning, and Key Frame Update steps.

\subsection{Key Frame Selection}
\label{ssec:key_frame_selection}

Key frame selection remains a topic worthy of research in recent years \cite{yoon2023exploring}. To minimize manual annotation, our approach involves selecting key frames from a sequence of continuous frames $X$. We employ the Structural Similarity Index (SSIM) as the metric to assess the distance between frames. Our algorithm, therefore, initializes the first frame as the current key frame $X[k_{c}]$ and sequentially evaluates the SSIM value between input frames $X[t]$ and the key frame $X[k_{c}]$ . Frames with an SSIM value higher than a threshold $s$ are added to a non-key frame sequence $X_{NK}$, and those below $s$ are added to the key frame sequence $X_{K}$, resetting the index of current key frame $k_{c}$. This method efficiently identifies significantly different frames as key frames, where the $s$ value impacts the number of images requiring labeling.
After key frame selection, we get $X = X_{K} + X_{NK}$ where $X_{K} = \{x_{k_{n}} \vert k_{n} \in \mathcal{K}\}$ is the key frames requiring manual annotation, $Y_{K} = \{y_{k_{n}} \vert k_{n} \in \mathcal{K}\}$ denoting the manually annotated labels, $X_{NK} = \{x_{k_{n}+ \Delta_{(n,i)} } \vert k_{n} \in \mathcal{K}, \Delta_{n}=[1, k_{n+1}-k_{n}-1]\}$ is non-key frame without annotations and K is index subset of the selected key frames $\mathcal{K} = [k_{1}, k_{2}, k_{3},...]$.

\RestyleAlgo{ruled}

\SetKwComment{Comment}{/* }{ */}

\begin{algorithm}
\caption{Key frame selection}\label{alg:two}
\KwData{Input image array $X$, Key frame array $X_{K}$, Non-key frame array $X_{NK}$, Key frame index $K$, Current key $k_{c}$, Threshold value $s$}
\KwResult{$X_{K}, X_{NK}, K$}
$k_{c} \gets 1$\;
$X_{K}[1] \gets X[1]$ \;
$X_{NK}[1] \gets 0$ \;
\For{$t \gets 2$ to $Length(X)$}{
    \eIf{ $SSIM(X[k_{c}], X[t]) \geq t$}{
    $X_{K}[t] \gets 0$ \;
    $X_{NK}[t] \gets X[t]$ \;
    }
    {
    $X_{K}[t] \gets X[t]$ \;
    $X_{NK}[t] \gets 0$ \;
    $K.add(t)$ \;
    $k_{c}=t$ \;
     }
  
}
\Return  $X_{K}, X_{NK}, K$
\end{algorithm}

\subsection{Model Initialize}
\label{ssec:model_initialize}

Upon acquiring key frames and completing their manual annotation, our methodology involves utilizing these key frames to initialize training for memory-based Video Object Segmentation \cite{yan2023two} and segmentation models \cite{chen2018encoder}. This process results in the derivation of two functions, $\mathcal{F}_{seg}$ and $\mathcal{F}_{vos}$, representing the segmentation model and the VOS model, respectively. In this framework, $\hat{y}_{t}$ denotes the inference made by the segmentation model, while $\hat{y}_{k_{n} \rightarrow t}$ represents the inference conducted by the VOS based on reference frames/labels pair $(x_{k_{n}}, y_{k_{n}})$ within the memory sequence. See Equation \ref{eq:1} and Equation \ref{eq:2}.

\begin{equation}\label{eq:1}
    \hat{y}_{t} = \mathcal{F}_{seg}({x}_{t}, \theta_{seg})\in{I}^{H \times W \times C} 
\end{equation}
\begin{equation}\label{eq:2}
     \hat{y}_{k_{n} \rightarrow t} = \mathcal{F}_{vos}(x_{t}, x_{k_{n}}, y_{k_{n}}, \theta_{vos})\in{I}^{H \times W \times C}
\end{equation}



In training phase, we exclusively utilize key frames and \textbf{only} their subsequent frames for training purposes. It ensures obtained $\hat{y}_{k_{n} \rightarrow k_{n}+1}$ based on previous frame. Following this, the key frames are updated based on the Temporal Consistent Loss, which is elaborated upon in the latter part of this paper. 

\subsection{Temporal Consistent Learning}
\label{ssec:TC_frame}

In this session, the key concept is leveraging temporal consistency loss to induce the segmentation model to depend on VOS for stable inference. This approach involves updating frames with stable inference into the key frame set, thereby allowing the introduction of new objects in the VOS's key frames. This strategy aims to enhance the adaptability and accuracy of the segmentation model by incorporating dynamic elements from VOS learning.

\begin{equation}\label{eq:3}
     f_{y_{i}}^{*} =[ v^{*Q}_{i}, \cfrac{1}{Z} \sum_{\forall j} \mathcal{f}(k^{*Q}_{i}, k^{*M}_{j}), v^{*M}_{j} ]
\end{equation}
\begin{equation}\label{eq:4}
     \mathcal{J}^{TC}_{k_{n} \rightarrow t} = \cfrac{1}{\mathcal{I}}\sum_{i \in \mathcal{I}} \Vert \hat{y}_{k_{n} \rightarrow t} , \hat{y}_{t} \Vert_{2}^{2}
\end{equation}

Equation \ref{eq:3} illustrates the inference mechanism of memory-based VOS. In the preceding phase, we incorporate the key frame into the memory frame, rendering the current VOS inference result as $\hat{y}_{k_{n} \rightarrow t}$, which denotes the outcome derived using key frame $k_{n}$ and current frame $t$. However, in adherence to the definition of temporal consistency loss \cite{varghese2021unsupervised}, each frame must be compared against the variations from its preceding frame. To accomplish this, we have modified the memory mechanism of VOS by assigning all key frames prior to the current frame into memory. Additionally, we have developed a key frame update algorithm, employing the Temporal Consistency (TC) loss to assess the consistency between segmentation and VOS inference. When the loss falls below a certain threshold, the current frame with inference from segmentation $\hat{y}_{t}$ is updated into the key frame sequence. By exclusively utilizing the subsequent frame of a key frame for training, it is ensured that the index $t-1$ is always within $k_{n}$. See as in algorithm \ref{alg:update}
.
\RestyleAlgo{ruled}

\SetKwComment{Comment}{/* }{ */}

\begin{algorithm}[htb]
\caption{Temporal Consistency Training and Key frame updating}\label{alg:update}
\KwData{Input image array $X$, Key frame array $X_{K}$, Key frame mask array $Y_{K}$, Temporal frame array $X_{T}$, Temporal mask array $\hat{Y}_{T}$, $\hat{Y}_{K \rightarrow T}$, Key frame index $K$, Threshold value $tc_{s}$}
\KwResult{$X_{T}$,$Y_{T}$,$X_{K}$,$Y_{K}$}

$X_{T} \gets$ all zeros array with $Length(X)$ \;
\For{$i \gets 1$ to $Length(K)$}{
    \If{$X_{T}[K[i]+1] = 0$}{
    $X_{T}[K[i]+1] \gets X[K[i]+1]$ \;
    }
}

\While{Train and Update}{
$\hat{Y}_{K} \gets \mathcal{F}_{seg}(X_{K}, \theta_{seg})$ \;
$\hat{Y}_{K \rightarrow T} \gets \mathcal{F}_{vos}(X_{T}, X_{K}, Y_{K},\theta_{vos})$ \;
$\hat{Y}_{T} \gets \mathcal{F}_{seg}(X_{T}, \theta_{seg})$ \;
Use Eq. \ref{eq:loss} to train $\mathcal{F}_{seg}$ \;
$\hat{Y}_{T} \gets \mathcal{F}_{seg}(X_{T}, \theta_{seg}^{'})$ \;
\For{$i \gets 1$ to $Length(\hat{Y}_{T})$}{
    \If{mIoU( $\hat{Y}_{K \rightarrow T}[i]$, $\hat{Y}_{T}[i]$ ) $> tc_{s}$}{
        $Y_{K}[i] \gets \hat{Y}_{T}[i]$ \; 
        $X_{K}[i] \gets X[i]$ \;
       \If{$X_{T}[i+1] = 0$}{
        $X_{T}[i+1] \gets X[i+1]$ \;
        }

        $K.insert(i)$

    }
}
}
\Return  $X_{T}$,$Y_{T}$,$X_{K}$,$Y_{K}$
\end{algorithm}

\subsection{Loss function}
\label{ssec:Loss_function}

\begin{equation}\label{eq:lkf}
     \mathcal{J}^{KF}_{k_{n}} = \cfrac{1}{\mathcal{I}}\sum_{i \in \mathcal{I}} \Vert  \hat{y}_{k_{n}}, y_{k_{n}}  \Vert_{2}^{2}
\end{equation}
\begin{equation}\label{eq:loss}
     \mathcal{J} = \alpha \times \mathcal{J}^{KF}_{k_{n}} + (1-\alpha) \times \mathcal{J}^{TC}_{k_{n} \rightarrow t}
\end{equation}

Following the implementation of TC loss and key frame update mechanisms, our method calculates key frame loss using frames updated into key frames. For frames not updated into key frames, we compute TC loss. The total loss is an aggregated weighted sum of these two losses.

\begin{figure}[htb]
    \centering
    \includegraphics[width=0.9\linewidth]{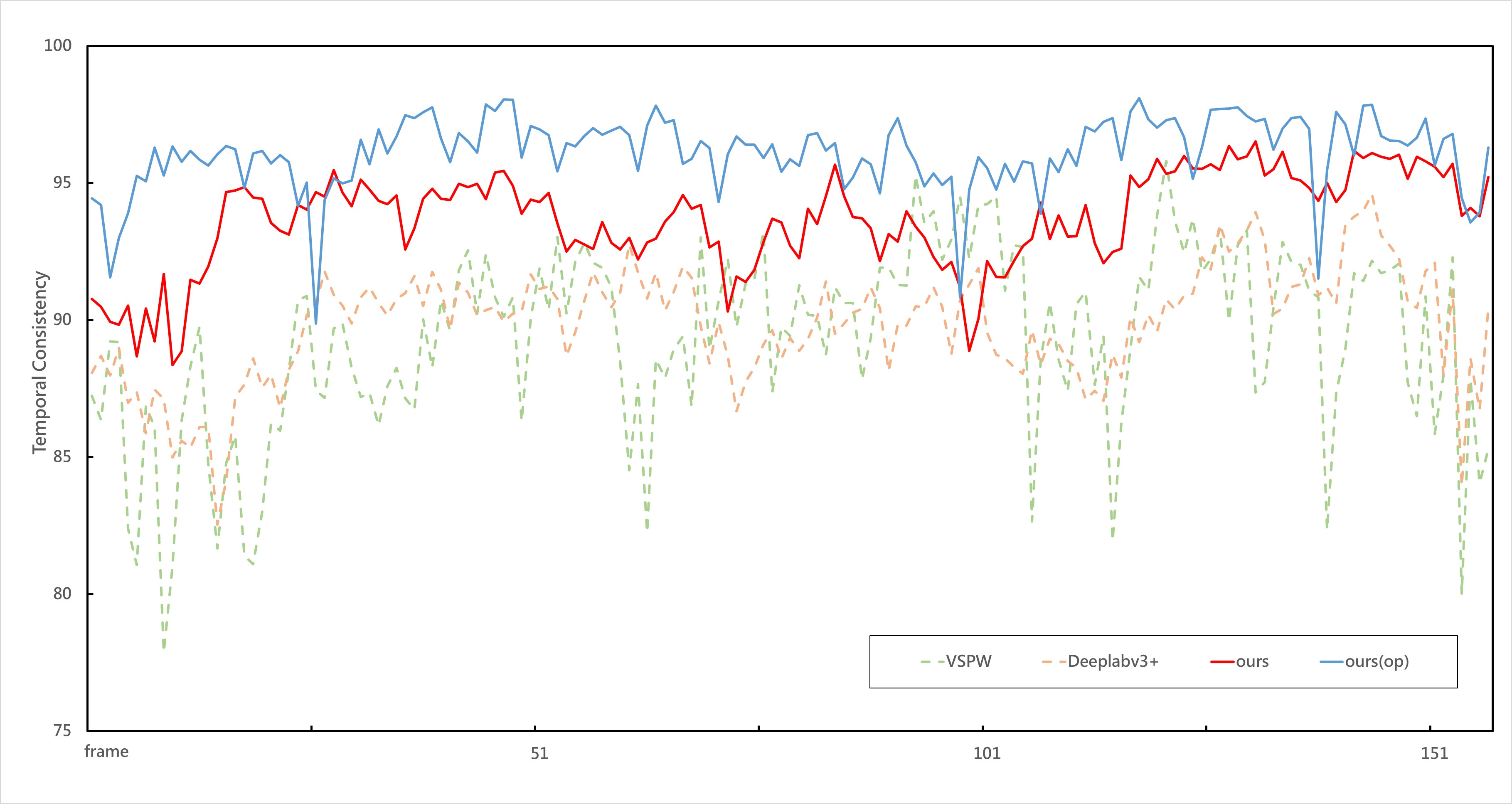}
    \caption{This fighure illustrates the temporal consistency of four models tested on a set of 158 validation images. In this depiction, the model developed under our architecture demonstrates better and more stable performance in comparison to the others.}
    \label{fig:TC_curve}
\end{figure}

\begin{figure*}[htb]
    \centering
    \includegraphics[width=0.95\linewidth]{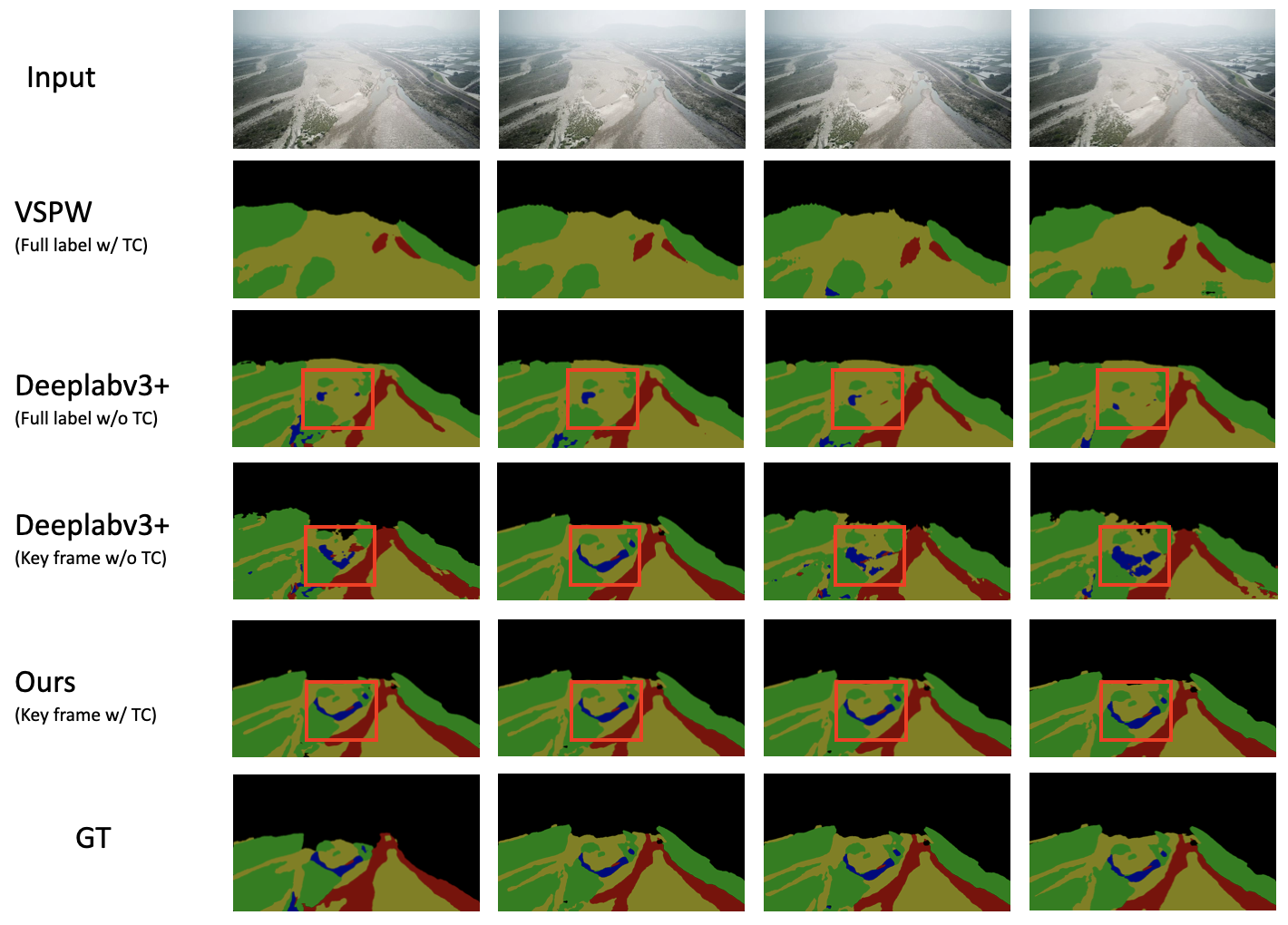}
    \caption{This figure demonstrates that using the entire labeled dataset for training does not necessarily enhance the accuracy of segmentation due to potential redundant training.The red box highlights the importance of TC in localization tasks. Our optimized method, however, is capable of improving both accuracy and temporal consistency under weakly supervised conditions.}
    \label{fig:result}
\end{figure*}

\section{Experiment results}
\label{sec:typestyle}

To evaluate the efficacy of our method, particularly in balancing the quantity of annotations against accuracy in few-shot learning, we extracted one image per second from a stream of UAV flight footage over rivers. From this, we selected 24 segments of flight imagery, totaling 1,553 images for training(full manual annotated for experiment). A subset of these annotated images, comprising 158 frames, was then chosen as the validation dataset. The dataset was captured during the dry season of the Dajia River and Da'an River in Taiwan.
In this experiment, we selected Two-shot VOS\cite{yan2023two} as the teacher and DeepLabv3+\cite{chen2018encoder} as the student model.

\subsection{Key Frame Density}
\label{ssec:key_frame_density}

Table 1 shows the number of annotations selected by Algorithm 1 at various SSIM threshold values. In our experiments, using thresholds of 0.5 and above consistently yielded mean Intersection over Union (mIoU) scores above 70\%, with the required amount of annotations being only about 20\% to 30\% of the training dataset.

\begin{table}[ht]\label{tab:ssim_kf}
\centering
\caption{The quantity of key frame sampling.}
\scalebox{0.85}{
\begin{tabular}{crrrrr}
\toprule
 
SSIM threshold  &  0.3 &  0.4 &  0.5 & 0.6 &  0.7    \\
\midrule
Number of key frames  & 152  & 223 & 324 &  455  & 679 \\
Annotation ratio   & 0.1 & 0.14 & 0.21 & 0.30 &  0.43  \\

\bottomrule
\end{tabular}
}

\end{table}

\subsection{Evaluation and Temporal Consistency}
\label{ssec:temporal_consistency}

Table 2 and Figure \ref{fig:TC_curve} illustrate that our method significantly improves temporal consistency. Notably, techniques based on optical flow or consecutive frames underperform in aerial views even trained with fully labeled segmentation training.
TC values also cannot be measured.
We attribute this to the lack of training data specific to optical flow networks. 
In the case of VSPW, after more than 20,000 training iterations, there was no improvement observed in the mIoU, which averaged at 54.52. Figure \ref{fig:result} shows the inference results of VSPW, where the sediment (blue area) could not be located.
Upon training with fully labeled data, DeepLabV3+ achieved mIoU of 73.4 and TC value of 89.8.
Our approach, selecting SSIM values of 0.5 and 0.6, outperforms fully labeled segmentation. In the case of SSIM 0.6, we achieved mIoU of 76.90 and TC of 95.00, utilizing only 30\% of the labeled data.

\begin{table}[ht]\label{tab:mIoU_TC}
\centering
\caption{The measurement results of mIoU and Temporal consistency}
\scalebox{0.70}{
\begin{tabular}{crrrrr}
\toprule
Approach  & Key frame &  Annotation & w/ TC loss & mIoU &  TC \\
\midrule
ETC VideoSeg\cite{liu2020efficient} & Full & 1553 & \Checkmark & 18.32 & \XSolidBrush \\
VSPW\cite{miao2021vspw} & Full & 1553 & \Checkmark & 54.52 & 89.40 \\
PSPnet & Full & 1553 & \XSolidBrush & 71.82 & 89.30\\
HRnet & Full & 1553 & \XSolidBrush & 72.63 & 90.50\\
Deeplabv3+\cite{chen2018encoder} & Full & 1553 & \XSolidBrush & 73.40 & 89.80 \\
\midrule
Deeplabv3+ $_{SSIM@0.3}$ & SSIM & 152 &\XSolidBrush & 70.94  & 88.80  \\
Deeplabv3+ $_{SSIM@0.4}$ & SSIM & 223 &\XSolidBrush & 73.41  & 89.30  \\
Deeplabv3+ $_{SSIM@0.5}$ & SSIM & 324 &\XSolidBrush & 74.00  & 88.90  \\
Deeplabv3+ $_{SSIM@0.6}$ & SSIM & 455 &\XSolidBrush & 74.13  &  89.80 \\
\midrule
Ours $_{avgselect}$ & Uniform & 152 & \Checkmark & 71.93  & 90.07  \\
Ours $_{avgselect}$ & Uniform  & 223 & \Checkmark & 72.33  & 91.50  \\
Ours $_{avgselect}$ & Uniform  & 324 & \Checkmark & 73.33  & 93.07  \\
Ours $_{avgselect}$ & Uniform   & 455 & \Checkmark & 73.51  & 93.51  \\
\midrule
Ours $_{SSIM@0.3}$ & SSIM & 152 & \Checkmark & 74.50  & 90.80  \\
Ours $_{SSIM@0.4}$ & SSIM & 223 & \Checkmark & 73.45  & 92.00  \\
Ours $_{SSIM@0.5}$ & SSIM & 324 & \Checkmark & 74.50  & 93.30  \\
Ours $_{SSIM@0.6}$ & SSIM  & 455 & \Checkmark & \textbf{76.90}  &  \textbf{95.00} \\
\bottomrule
\end{tabular}
}

\end{table}

\subsection{Ablation Study}
\label{ssec:ablation_study}

In Table 2, we tested key frame selections at SSIM thresholds from 0.3 to 0.6 resulting in the extraction from 152 to 455 frames respectively.
We compared the performance of models under four training scenarios: 1. Standard training, involving the annotation of all data followed by training. 2. Selection of key frames for training, where only key frames are used for standard training. 3. Within our proposed architecture, key frames are selected through average sampling and subjected to temporal consistency training. 4. In our architecture, key frames are chosen based on the SSIM and undergo temporal consistency training.

In comparing the performance of models under conventional training with those subjected to alternative training scenarios, we observe that the utilization of key frames may indeed surpass the results attained with fully labeled datasets. This superiority could be attributed to the reduction in redundant training afforded by the strategic selection of key frames.
However, as distinctly observed in Figure \ref{fig:result}, models boasting higher mIoU may still falter in aerial tasks due to a decline in TC, leading to failures in localization tasks. For instance, the instability in locating sand sources, as highlighted by the red box, could severely impact the analytical outcomes.

Models trained solely on key frames may surpass the performance metrics obtained with fully labeled datasets in terms of mIoU. However, as clearly depicted in Figure 4, solely enhancing mIoU without sufficient maintenance of TC can lead to failures in accurately locating sand sources. Our approach, at an SSIM value of 0.4 and utilizing only 14\% of the data, achieves superior results compared to using fully labeled datasets. When the SSIM threshold is set to 0.6, there is an improvement of 3.5\% in mIoU and a 5.2\% increase in TC, allowing for consistent tracking and localization of exposed land and sediment across multiple instances.

\section{Conclusion}
\label{sec:copyright}

In this study, we introduce a teacher-student model composed of Video Object Segmentation (VOS) and segmentation. We developed key frame selection and update mechanisms to achieve weakly supervised learning and distilled temporal consistency knowledge. The experimental results show that our method enhances both mIoU and temporal consistency using fewer annotated data. This approach overcomes traditional limitations of temporal consistency in aerial scenarios, advancing environmental engineering tasks in UAV river monitoring for sediment and exposed land localization.

\vfill\pagebreak



\bibliographystyle{IEEEbib}
\bibliography{strings,refs}

\end{document}